# Mechanics of fiber reinforced soft manipulators based on inhomogeneous Cosserat rod theory

Sadegh Pourghasemi Hanza[a], Hamed Ghafarirad[a*]

[a] *Department of Mechanical Engineering, Amirkabir University of Technology, Tehran, Iran*

*Hamed Ghafarirad, Department of Mechanical Engineering, Amirkabir University of Technology, 424, Hafez Avenue, Tehran 15875-4413, Iran.
Email: Ghafarirad@aut.ac.ir

# Mechanics of fiber reinforced soft manipulators based on inhomogeneous Cosserat rod theory

In this study, an inhomogeneous Cosserat rod theory is introduced and compared to the conventional homogeneous rod for modeling soft manipulators. The inhomogeneity is addressed by considering the pressure actuation as part of the rod's constitutive laws, resulting in shifting the neutral axis. This shift is investigated for a soft manipulator with three parallel fiber-reinforced actuators. Furthermore, a fiber-reinforced actuator is modeled using nonlinear continuum mechanics to extract the effect of radial pressure on axial deformation and is combined with Cosserat model. Finally, several numerical methods are employed to solve the proposed model and validated by a series of experiments.

## 1. Introduction

Nowadays, soft robots as a branch of continuous robots have extensively attracted the attention of researchers. Due to their soft structure, these robots are ideal for tasks such as safe interaction, complex deformations, and impact resistance [1].

According to different applications, they have been produced with various designs and methods. Some are made using electroactive polymers [2, 3], shape-memory alloys [4, 5] and piezoelectric actuators [6, 7]. But pneumatically actuated soft actuators are the most commonly used. Examples include the McKibben actuator [8], muscle motor actuator [9], pouch motors [10], flexible microactuator [11], PneuNet actuator [12] and fiber-reinforced bending actuator [13, 14].

Recently, interests in the modeling of fiber-reinforced actuators have been increased. Bishop-Moser et al. patented generalized fiber-reinforced actuators consisting of a tube encircled by two families of fibers [15]. Numerous techniques have been explored to model deformation and interaction response of these actuators, but it is a complex problem due to factors like large deformations, actuator compliance, and nonlinearity of materials. While finite element analysis produces the most precise results [16, 17], it is extremely time-consuming and unsuitable for real-time implementations. On the other hand, analytical modeling demands a careful selection of assumptions and approximations in order to provide tractable and accurate results [18, 19].

However, since each actuator has just one internal chamber, it can only create one sort of motion. Hirai et al. [20] created a spatial manipulator by combining multiple actuators in parallel and avoided this limitation. This type of manipulator is simulated using lumped-mass models [21, 22] and finite element models [23]. These methods provide models with many degrees of freedom, making control design difficult. On the other hand, constant curvature technique produces models with fewer degrees of freedom and has been used to simulate and control soft continuum robots [24, 25]. These models need considerable identification and typically only produce consistent results for a small portion of the robot workspace and are only valid in the absence of external forces.

It is possible to generate models with the predictive accuracy of the finite element approach while maintaining a limited degree of freedom using rod theories [26, 27]. Typically, Cosserat rods with a linear elastic material law are utilized, either spatially discretized using finite element methods [28, 29] or the shooting method [30]. On the other hand, due to geometrical factors such as finite displacements and rotations as well as contact and friction forces, a nonlinear Cosserat theory is more accurate [31] and can be applied to a continuum manipulator, by including chamber pressurization into constitutive law [32].

In this study, a soft spatial manipulator consisting of three pneumatic fiber-reinforced actuators is introduced. The shape of each actuator is affected by the internal air pressure in two ways. First, the air pressure acts directly to produce an internal force along the axial direction of the actuator. Second, the air pressure acts radially, and the fiber reinforcement causes this radial pressure to couple to additional axial stress due to the relative inextensibility of the fibers. This issue is well studied in [33] from a continuum mechanical point of view, assuming the inextensibility of fibers. Furthermore, the extensibility of fibers is considered in [34] using a nonlinear continuum mechanics approach. In this paper, a similar approach is implemented to model each fiber-reinforced actuator. This model is used to calculate the radial pressure effect (RPE) on the axial deformation of each actuator. Later, this effect will be combined with the dynamical equations of the spatial manipulator. To extract these equations, two cases of homogeneous and inhomogeneous Cosserat rods are investigated. These equations are then numerically solved to validate the experimental results. Here, since general closed-form solutions are not possible, a variety of numerical methods are employed and compared in terms of accuracy and computation performance.

The remainder of this paper is organized in the following manner: In Section 2, the manufacturing procedure for the spatial soft manipulator is introduced. In Section 3, the RPE on the axial deformation is calculated using a continuum-based model for a single fiber-reinforced actuator. In Section 4, a Cosserat theory-based model of the manipulator (proposed in Section 2) is derived and the axial deformation resulting from RPE (calculated in Section 3) is added to it. In Section 5, the unknown parameters are identified using experimental tests in order to simulate the proposed model. In Section 6, the experimental and simulation results are compared, and finally, conclusions are presented.

## 2. Soft pneumatic manipulator

A soft continuum manipulator consists of three fiber-reinforced actuators, arranged symmetrically around the manipulator vertical axis and connected to each other by two 3D-printed rigid caps made of PLA at both ends and eight 3D-printed flexible parts made of TPU in the middle. Each actuator is a cylindrical elastomeric tube and is molded in multiple stages with 3D-printed casting molds. Two fibers arranged in a helical pattern are wound over the tube in order to decrease its radial expansion and force the actuator to enlarge only in the longitudinal direction. A thin silicon layer covers the fibers to keep them from moving during actuation. (Figure 1)

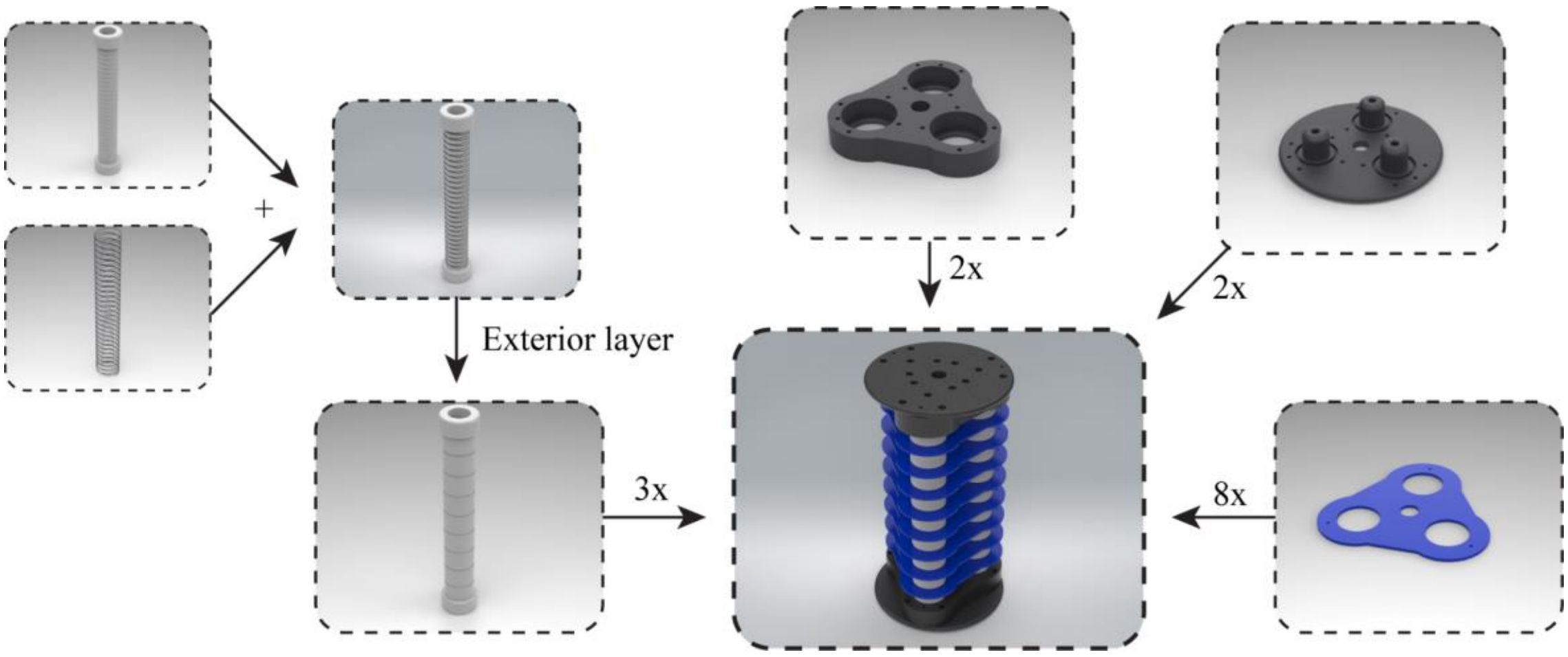

Figure 1. Different parts of the proposed manipulator

The end cap constrains the actuators in all directions. On the other hand, flexible parts (blue ones) connect the middle of actuators, allowing only axial deformation and constraining the other DOFs. This will prevent the actuators from buckling and create a continuous deformation for the manipulator. To support the reader's understanding, all geometric parameters are introduced in Figure 2.

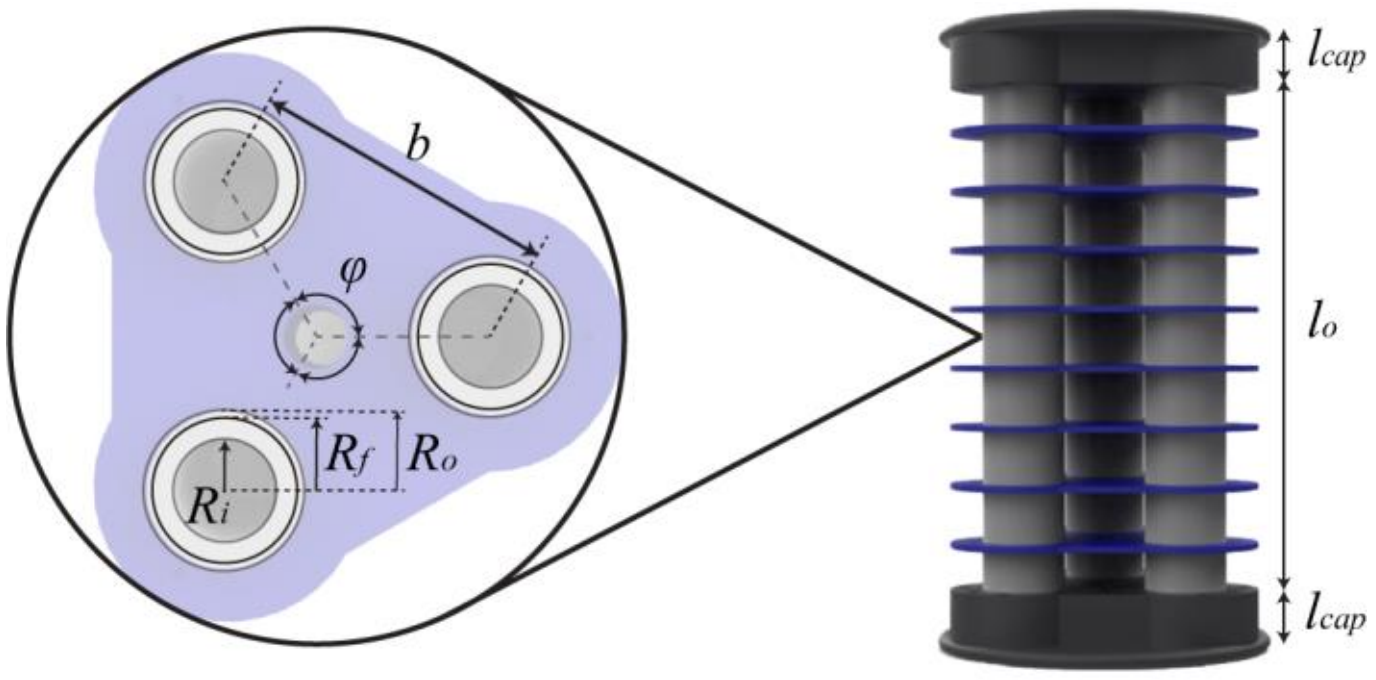

Figure 2. Geometric parameters of the manipulator

The manipulator parameters and their values are listed in Table 1.

Table 1. Geometrical parameters of the manipulator

| **Parameter** | **Value** |
|---|---|
| $b$ | 55 mm |
| $R_i$ | 9.5 mm |
| $R_f$ | 12.5 mm |
| $R_o$ | 14 mm |
| $l_o$ | 170 mm |
| $l_{cap}$ | 15 mm |
| $\varphi$ | 120° |

## 3. Single-actuator model

A fiber-reinforced actuator may perform a wide variety of motions depending on the fiber angle and material used. When the elastomeric tube is homogenous and symmetrically reinforced with fibers, it undergoes some combination of axial extension and radial expansion upon pressurization. This pressure exerts some force radially on the walls of the cylinder and axially on the end cap. It can be shown that not only the axial pressure but also the radial pressure has some role in the axial extension.

Here, the effect of radial pressure is modeled using a nonlinear elasticity method. The actuator is considered to be a hollow cylinder of isotropic incompressible hyperelastic material surrounded by a thin layer of anisotropic material (fibers and a thin layer of silicon) with continuous deformation between the two layers. Initially, the isotropic core has an inner radius of $R_i$ and an outer radius of $R_m$, whereas the anisotropic layer has an outer radius of $R_o$. The direction of the anisotropic material with a fiber angle of $\Psi$ is considered by the initial fiber orientation $\boldsymbol{S} = (0, \cos\Psi, \sin\Psi)$. It is assumed that when the tube is pressurized, it preserves its cylindrical shape and the radii become $r_i$, $r_m$ and $r_o$. The possible extension and expansion deformations are then described by Eq. (1).

$$\boldsymbol{F} = \begin{bmatrix} \frac{\partial r}{\partial R} & \frac{1}{R}\frac{\partial r}{\partial \Theta} & \frac{\partial r}{\partial Z} \\ r\frac{\partial \theta}{\partial R} & \frac{r}{R}\frac{\partial \theta}{\partial \Theta} & r\frac{\partial \theta}{\partial Z} \\ \frac{\partial z}{\partial R} & \frac{1}{R}\frac{\partial z}{\partial \Theta} & \frac{\partial z}{\partial Z} \end{bmatrix} = \begin{bmatrix} \frac{R}{r\lambda_z} & 0 & 0 \\ 0 & \frac{R}{r} & 0 \\ 0 & 0 & \lambda_z \end{bmatrix} \tag{1}$$

where $R, \Theta, Z$ and $r, \theta, z$ are the radial, circumferential and longitudinal coordinates in the reference and current configurations, respectively and $\lambda_z$ denotes the axial stretch (Figure 3).

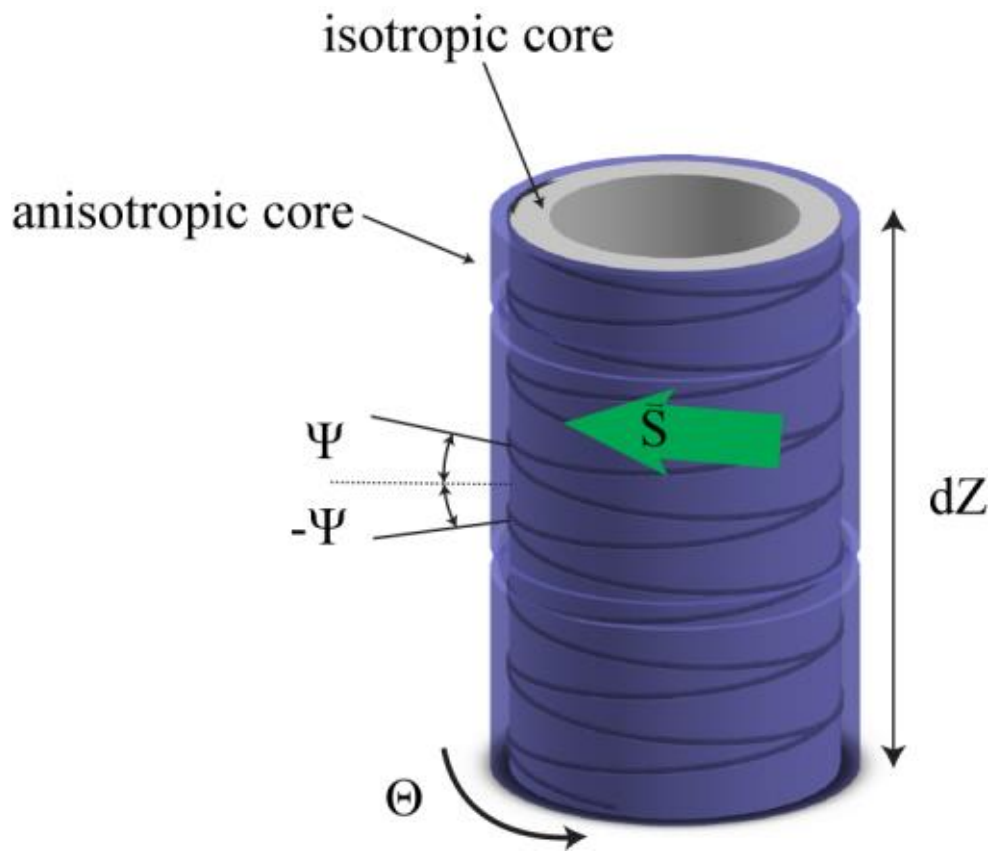

Figure 3. Required parameters to model a fiber reinforced actuator

For the isotropic core, a simple incompressible Neo-Hookean model is adopted, in which strain energy is denoted by $W^{(iso)} = \mu/2(I_1 - 3)$, where $\mu$ is the initial shear modulus and $I_1 = tr(\boldsymbol{FF}^T)$. For the anisotropic layer, strain energy is considered as

$$W^{(out)} = c_1 W^{(iso)} + c_2 W^{(aniso)} \tag{2}$$

where $W^{(iso)}$ and $W^{(aniso)}$ denote the contribution of the isotropic and anisotropic sections and $c_i$ is the volume fraction of each section. By modeling the fiber as a rod subjected to an axial load, the strain energy density is calculated as $W^{(aniso)} = \frac{(\sqrt{I_4}-1)^2 E}{2} + \frac{(\sqrt{I_6}-1)^2 E}{2}$, where $E$ is its Young's modulus, $I_4 = \boldsymbol{s}_1.\boldsymbol{s}_1$, $I_6 = \boldsymbol{s}_2.\boldsymbol{s}_2$ and $\boldsymbol{s}_\alpha = \boldsymbol{FS}_\alpha$.

Then, by having strain energies, the Cauchy stresses can be derived as Eq. (3) where $W_i = \frac{\partial W}{\partial I_i}$, $\boldsymbol{I}$ is the identity matrix, and $p$ is the hydrostatic pressure.

$$\begin{gathered}\boldsymbol{\sigma}^{(in)} = 2W_1^{(in)}\boldsymbol{FF}^T - p\boldsymbol{I} \\ \boldsymbol{\sigma}^{(out)} = 2W_1^{(out)}\boldsymbol{FF}^T + 2W_4^{(out)}\boldsymbol{s}_1 \otimes \boldsymbol{s}_1 + 2W_6^{(out)}\boldsymbol{s}_2 \otimes \boldsymbol{s}_2 - p\boldsymbol{I}\end{gathered} \tag{3}$$

The Cauchy equilibrium equations for the radial direction result in $\frac{d\sigma_{rr}}{dr} = \frac{\sigma_{\theta\theta}-\sigma_{rr}}{r}$, which can then be integrated to provide Eq. (4) for the radial stress.

$$\Delta\sigma_{rr} = \int_{r_i}^{r_m} \frac{\sigma_{\theta\theta}^{(in)} - \sigma_{rr}^{(in)}}{r}\, \mathrm{d}r + \int_{r_m}^{r_o} \frac{\sigma_{\theta\theta}^{(out)} - \sigma_{rr}^{(out)}}{r}\, \mathrm{d}r \tag{4}$$

The axial load $N$ can also be obtained as:

$$N = 2\pi \int_{r_i}^{r_m} \sigma_{zz}^{(in)} r\, \mathrm{d}r + 2\pi \int_{r_m}^{r_o} \sigma_{zz}^{(out)} r\, \mathrm{d}r \tag{5}$$

Assuming the pressure $P$ is applied to the actuator and there is no external force or moment, radial stress and axial force become $\Delta\sigma_{rr} = P,\ N = P\pi r_i^2$ and Eqs. **Error! Reference source not found.** and (5) can be solved to find axial stretch $\lambda_z$ and radial stretch $\frac{r_i}{R_i}$. This condition is well addressed in [34], which numerically solves relevant equations using Taylor expansion. The same approach can be used to solve the equations when the actuator is only under external force without any pressure, which is $\Delta\sigma_{rr} = 0$, $N = F_{ext}$. These two conditions can be implemented experimentally to identify the unknown parameters of the resulting Taylor equations. When the pressure is applied only in radial direction $\Delta\sigma_{rr} = P$, with no effect in the axial direction $N = 0$, the axial stretch $\lambda_z$ and radial stretch $\frac{r_i}{R_i}$ can be calculated theoretically as functions of $P$.

$$\lambda_z = f(P), \ \frac{r_i}{R_i} = g(P) \tag{6}$$

These stretches will be applied as RPE to Cosserat equations in the next section.

## 4. Manipulator model

In this section, a dynamical model for the continuum manipulator, containing three actuators at a distance of $\boldsymbol{r}_i$ from the central axis, will be derived. Due to the existence of flexible constraints to connect the actuators along the robot, these vectors remain constant throughout the neutral axis ($\boldsymbol{r}_{si} = \boldsymbol{0}$). In the formulation, the continuum robot will be considered as a Cosserat rod with general external forces and moments [35]. Fluid pressure is applied to the actuators, which results in a bending motion of the robot. A quasi-static fluid dynamic is assumed in the actuators so that there is a single uniform pressure $P_i(t)$ for each actuator.

### *4.1 Rod kinematics*

A Cosserat rod is approximated as a one-dimensional curve in space where the position and orientation of each point $s$ could be addressed by a vector $\boldsymbol{p}(s,t)$ and an orthonormal triad $\boldsymbol{R}(s,t) = (\boldsymbol{d}_1, \boldsymbol{d}_2, \boldsymbol{d}_3)$, respectively (Figure 4). The cross section is spanned by $d_1$ and $d_2$, and $\boldsymbol{d}_3 = \boldsymbol{d}_1 \times \boldsymbol{d}_2$. Generalized curvature $\boldsymbol{u}(s,t)$ and linear strain $\boldsymbol{v}(s,t)$ can be calculated as the partial derivative of $\boldsymbol{p}$ and $\boldsymbol{R}$, respectively, with respect to arc length $s$ in the local frame.

$$\boldsymbol{p}_s = \boldsymbol{R}\boldsymbol{v} \tag{7}$$

$$\boldsymbol{R}_s = \boldsymbol{R}\hat{\boldsymbol{u}} \tag{8}$$

where $\hat{\boldsymbol{u}} = \begin{bmatrix} 0 & -u_3 & u_2 \\ u_3 & 0 & -u_1 \\ -u_2 & u_1 & 0 \end{bmatrix}$.

The generalized curvature $\boldsymbol{u}$ represents the bending rates $u_1, u_2$ about the principal directions $(\boldsymbol{d}_1, \boldsymbol{d}_2)$ and the twisting rate $u_3$ about $\boldsymbol{d}_3$ while $\boldsymbol{v}$ is associated with the shears $v_1, v_2$ and the axial strain $v_3$.

To avoid truncation errors from numerically integrating $\boldsymbol{R}_s$, the orientation in quaternion form is used [36]. In Eq. (9), the equivalent equation of the rotation matrix and its derivation are calculated by introducing $\boldsymbol{h} = h_1 + h_2 i + h_3 j + h_4 k$.

$$\boldsymbol{R}(\boldsymbol{h}) = \boldsymbol{I} + \frac{2}{\boldsymbol{h}^T\boldsymbol{h}} \begin{bmatrix} -h_3^2 - h_4^2 & h_2h_3 - h_4h_1 & h_2h_4 + h_3h_1 \\ h_2h_3 + h_4h_1 & -h_2^2 - h_4^2 & h_3h_4 - h_2h_1 \\ h_2h_4 - h_3h_1 & h_3h_4 + h_2h_1 & -h_2^2 - h_3^2 \end{bmatrix}$$

$$\boldsymbol{h}_s = \frac{1}{2} \begin{bmatrix} 0 & -u_1 & -u_2 & -u_3 \\ u_1 & 0 & u_3 & -u_2 \\ u_2 & -u_3 & 0 & u_1 \\ u_3 & u_2 & -u_1 & 0 \end{bmatrix} \begin{bmatrix} h_1 \\ h_2 \\ h_3 \\ h_4 \end{bmatrix} \tag{9}$$

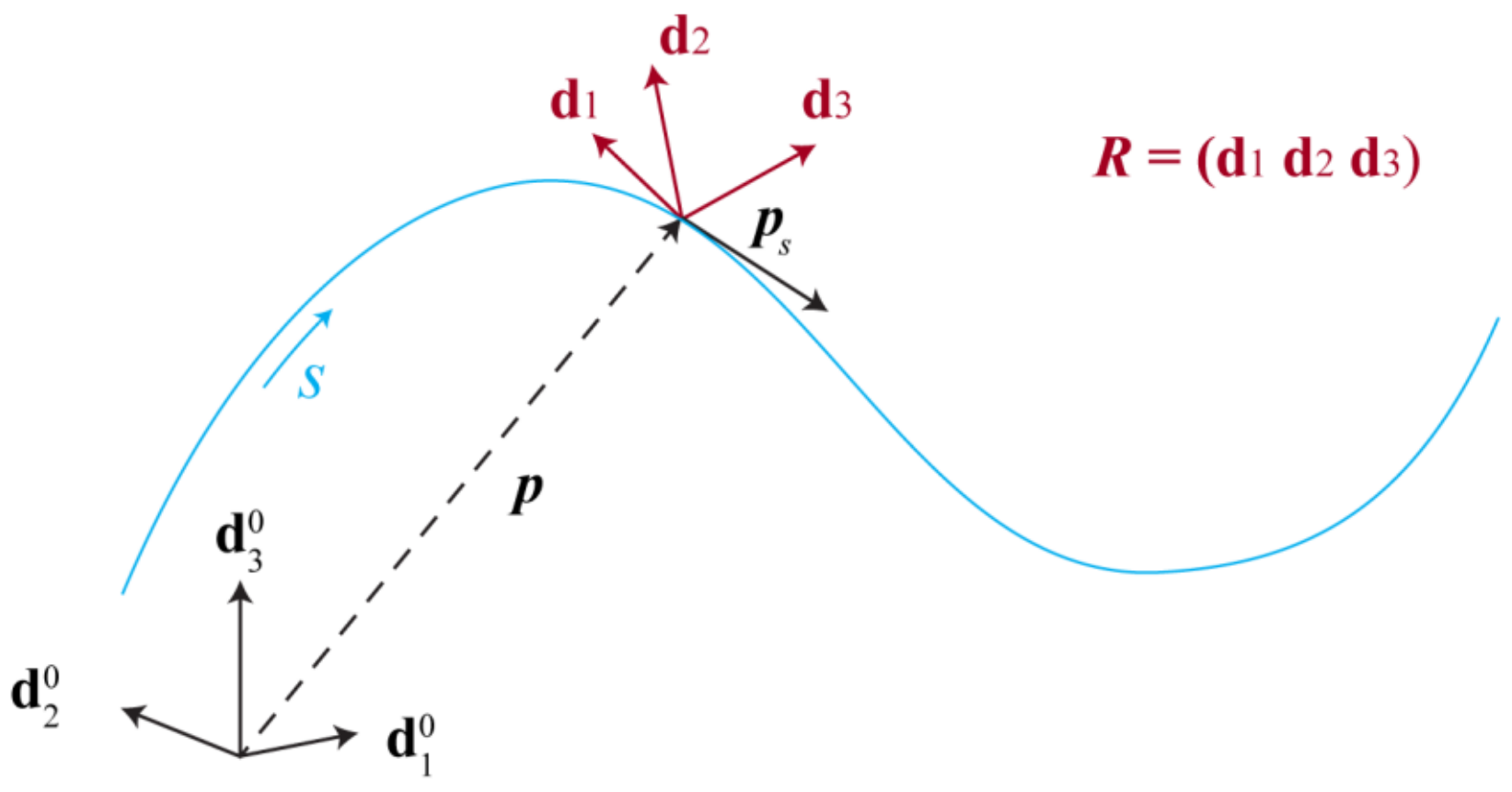

Figure 4. Kinematical parameters for Cosserat rod model

### *4.2. Compatibility laws*

Similar to $\boldsymbol{u}$ and $\boldsymbol{v}$, $\boldsymbol{\omega}$ and $\boldsymbol{q}$ represent angular and linear velocity for the rod, respectively, and are calculated in Eq. (10).

$$\begin{aligned} \boldsymbol{q}(t,s) &:= \boldsymbol{R}^T \boldsymbol{p}_t \\ \boldsymbol{\omega}(t,s) &:= (\boldsymbol{R}^T \boldsymbol{R}_t)^\vee \end{aligned} \tag{10}$$

Due to equality $\boldsymbol{p}_{st} = \boldsymbol{p}_{ts}$ and $\boldsymbol{R}_{st} = \boldsymbol{R}_{ts}$, by the combination of Eqs. (7) and (8), the spatial derivative of velocities can be derived as Eq. (11).

$$\begin{aligned} \boldsymbol{q}_s &= \boldsymbol{v}_t - \hat{\boldsymbol{u}}\boldsymbol{q} + \hat{\boldsymbol{\omega}}\boldsymbol{v} \\ \boldsymbol{\omega}_s &= \boldsymbol{u}_t - \hat{\boldsymbol{u}}\boldsymbol{\omega} \end{aligned} \tag{11}$$

### *4.3. Balance laws*

For an arbitrary section of the manipulator, as shown in Figure 5, the equilibrium of forces and moments leads to Eqs. (12) and (13) where $\rho$ is the mass density, $\boldsymbol{J}$ is the matrix of second area moments of the cross section, and $A$ is the area of the cross section.

$$\boldsymbol{n}_s = -\boldsymbol{f}_e + \rho A \boldsymbol{R}(\hat{\boldsymbol{\omega}}\boldsymbol{q} + \boldsymbol{q}_t) + \sum_{i=1}^{3} P_i A_i \boldsymbol{R}_s \boldsymbol{e}_3 \tag{12}$$

$$\boldsymbol{m}_s = -\boldsymbol{l}_e - \boldsymbol{p}_s \times \boldsymbol{n} + \partial_t(\boldsymbol{R}\rho\boldsymbol{J}\boldsymbol{\omega}) + \sum_{i=1}^{3} P_i A_i \boldsymbol{R}[(\boldsymbol{v} + \hat{\boldsymbol{u}}\boldsymbol{r}_i) \times \boldsymbol{e}_3 + \boldsymbol{r}_i \times \hat{\boldsymbol{u}}\boldsymbol{e}_3] \tag{13}$$

For infinitesimal $ds$, $\boldsymbol{n}$ and $\boldsymbol{m}$ represent the internal force and moment acting from the material at $\boldsymbol{p}(t, s + ds)$ on the material at $\boldsymbol{p}(t, s - ds)$, respectively, and are introduced in the global frame [30].

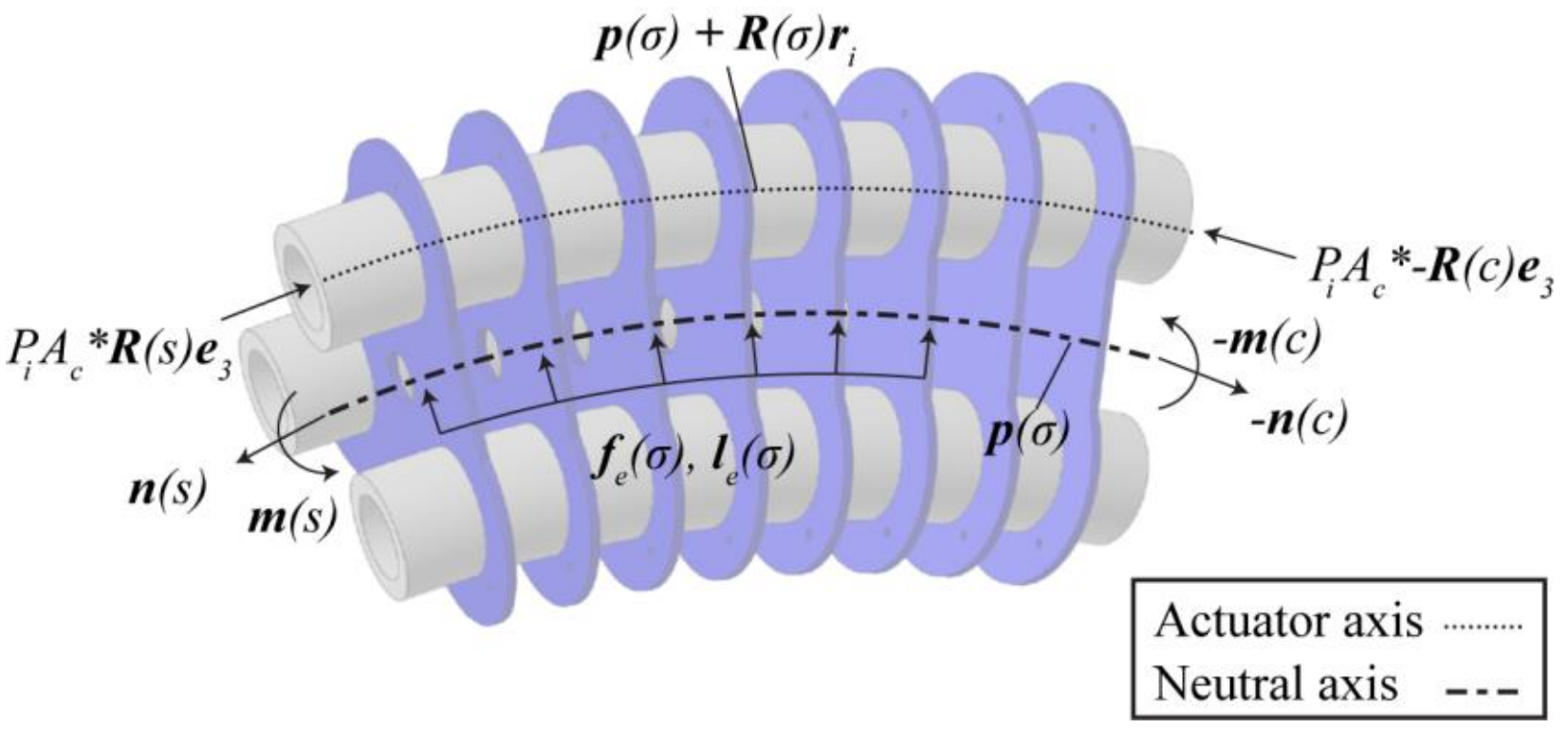

Figure 5. Acting forces and moments on a segment of the manipulator for cuts at $s$ and $c$ with $s > c$. $f_e$ and $l_e$ are the distributed external forces and moments, respectively.

### *4.4. Radial pressure effect (RPE)*

As mentioned in section 0, for a single actuator, radial pressure has an effect on both radial and axial strains. The change in radius doesn't affect Cosserat model, which is a one-dimensional rod, but the change in length is necessary to be considered. By Eq. (6), the axial strain for each individual actuator can be obtained as a function of internal pressure, where in terms of Cosserat strains, it becomes Eq. (14).

$$v_{3i}^{Aa} = f(P_i) \tag{14}$$

where $i$ is the number of actuators, superscript $Aa$ refers to the actuator axis and subscript 3 refers to $\boldsymbol{d}_3$ direction.

Despite the position of the neutral axis of the rod, these strains occur along the axis of each actuator and can be addressed in two ways.

- In the first approach, assuming that the cross-sectional area remains the same after deformation, the axial strain at each point of the cross-section is obtained from the relation $v_3^{Aa}(x, y) = v_3^{Na} + xu_1 + yu_2$ where $x$ and $y$ refer to the position of each actuator relative to the neutral axis. Therefore, the strains resulting from radial pressure along the actuator axis can be transferred to the neutral axis by equation $v_3^{Na} = v_3^{Aa}(x, y) - xu_1 - yu_2$.

- In the second approach, an equivalent pressure $P_i^r = \left(\frac{EA_m v_3^{Aa}}{A_{in}}\right)_i$ as a function of the strains is introduced, where $E$ is the Young's modulus obtained later in Eq. (17), $A_{in} = \pi R_i^2$ and $A_m = \pi(R_o^2 - R_i^2)$. This pressure can be added to the actual pressure and the new pressure becomes $P_i^{new} = P_i^{fluid} + P_i^r$. This pressure is later treated as real pressure and is used in dynamical equations.

### *4.5. Constitutive equations*

Appropriate constitutive relations are employed to relate the kinematic variables $\boldsymbol{u}$ and $\boldsymbol{v}$ to the internal forces $\boldsymbol{n}$ and moments $\boldsymbol{m}$ in the derived dynamical equation. Here, an elastic law is utilized as follows:

$$\begin{aligned} \boldsymbol{n} &= \boldsymbol{R}\boldsymbol{K}_{se}(\boldsymbol{v} - \boldsymbol{v}^*) \\ \boldsymbol{m} &= \boldsymbol{R}\boldsymbol{K}_{bt}(\boldsymbol{u} - \boldsymbol{u}^*) \end{aligned} \tag{15}$$

where "$se$" subscript refers to shear and extension, "$bt$" refers to bending and torsion, and "$*$" refers to initial condition.

The $\boldsymbol{K_{se}}$ and $\boldsymbol{K_{bt}}$ are stiffness coefficient matrices, which are determined by the material properties and cross-sectional geometry. The manipulator according to linearity or nonlinearity of actuators could be studied in two cases:

#### *4.5.1. Homogeneous cross-section*

For actuators with a linear strain-force relationship (constant Young's modulus), the manipulator consisting of three actuators is also linear and has a homogeneous cross-section. Therefore, the stiffness coefficient matrices can be extracted as:

$$\boldsymbol{K}_{se} = 3 * \begin{bmatrix} GA_m & 0 & 0 \\ 0 & GA_m & 0 \\ 0 & 0 & EA_m \end{bmatrix}, \qquad \boldsymbol{K}_{bt} = \begin{bmatrix} EI_{xx} & 0 & 0 \\ 0 & EI_{yy} & 0 \\ 0 & 0 & GI_{zz} \end{bmatrix} \tag{16}$$

$$G = \frac{E}{2f_s(1+\nu)}, \qquad I_{xx} = I_{yy} = 3I_o + \frac{A_m b^2}{2}, \qquad I_{zz} = I_{xx} + I_{yy}$$

where $\nu$ is Poisson ratio and $I_o = \frac{\pi(R_o^4 - R_i^4)}{4}$. The constant $f_s$ is form factor for shear and depends on cross sectional geometry [37]. Due to the complexity of the cross-section in proposed spatial manipulator, this constant will be identified experimentally.

### *4.5.2. Inhomogeneous cross-section*

For actuators with a nonlinear strain-force relationship, it can be concluded that the behavior of actuators depends on fluid pressure and external forces. As mentioned in section 0, the strains of a single actuator, in addition to the internal force and moment, are also a function of radial pressure $\boldsymbol{v} = \boldsymbol{v}(P_r, \boldsymbol{n}), \boldsymbol{u} = \boldsymbol{u}(P_r, \boldsymbol{m})$. On the other hand, internal loads are a function of axial pressure and external loads $\boldsymbol{n}(P_a, \boldsymbol{M}), \boldsymbol{m}(P_a, \boldsymbol{F})$. Therefore, nonlinearity can be addressed by a variable Young's modulus that is a function of pressure and external loads.

$$\boldsymbol{v} = \boldsymbol{E}(P, \boldsymbol{M})\boldsymbol{n}, \qquad \boldsymbol{u} = \boldsymbol{E}(P, \boldsymbol{F})\boldsymbol{m} \tag{17}$$

In other words, by applying different pressures and external forces to the actuators, the cross-section of the manipulator will have different Young's moduli and this will result in an inhomogeneous rod. The neutral axis of this inhomogeneous rod depends on the stiffness values of each actuator, and its position with respect to the central axis is calculated as Eq. (18) [38]. (Figure 6)

$$\bar{x}_{Na} = \frac{\sqrt{3}b}{6} * \frac{2E_1 - E_2 - E_3}{E_1 + E_2 + E_3}, \qquad \bar{y}_{Na} = \frac{b}{2} * \frac{E_2 - E_3}{E_1 + E_2 + E_3} \tag{18}$$

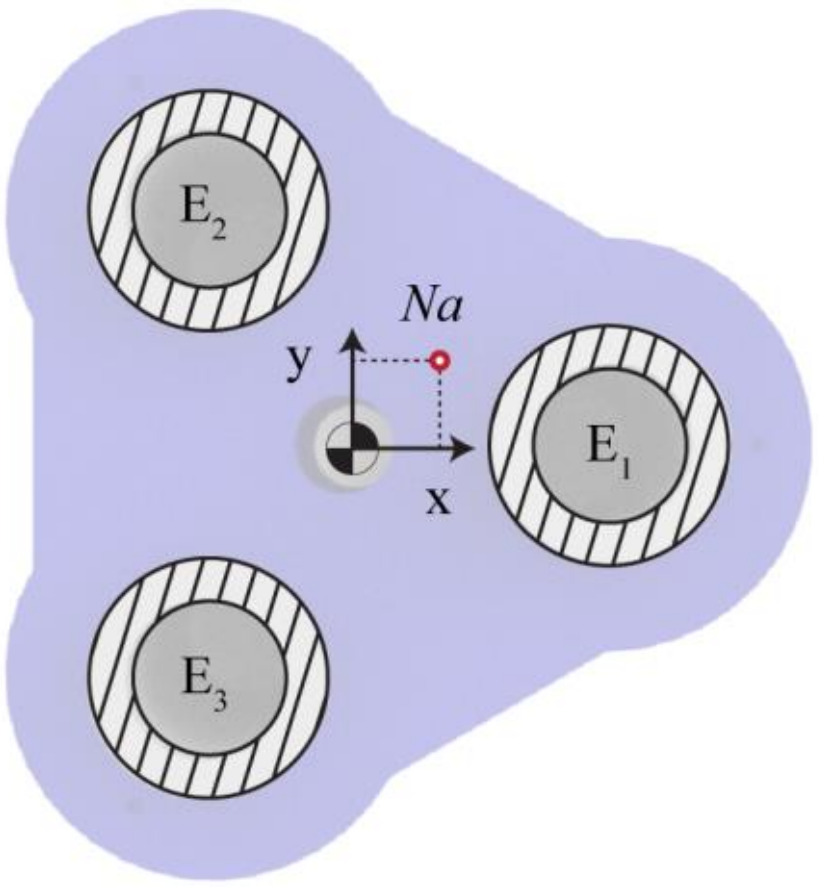

Figure 6. The Young's moduli of actuators and the neutral axis are not constant and will change according to the applied pressure and external loads.

This shift will change $\boldsymbol{r}_i$, and the new value is obtained in Eq. (19) by introducing $\boldsymbol{D}_{Na} = [\bar{x}_{Na} \quad \bar{y}_{Na} \quad 0]^T$.

$$\boldsymbol{r}_i^{new} = \boldsymbol{r}_i - \boldsymbol{D}_{Na} \tag{19}$$

By applying this displacement to the neutral axis, the stiffness matrix for longitudinal strains is defined as follows:

$$\boldsymbol{K}_{se} = \sum_{i=1}^{3} E_i * \begin{bmatrix} \frac{A_m}{2f_s(1+\nu)} & 0 & 0 \\ 0 & \frac{A_m}{2f_s(1+\nu)} & 0 \\ 0 & 0 & A_m \end{bmatrix} \tag{20}$$

On the other hand, the rotational stiffness matrix can be calculated as Eq. (21).

$$\boldsymbol{K}_{bt} = \begin{bmatrix} K_{11} & K_{12} & 0 \\ K_{12} & K_{22} & 0 \\ 0 & 0 & K_{33} \end{bmatrix}$$

$$K_{11} = EI_o + A_m \left( E_1 \bar{y}_{Na}^2 + E_2 \left( \frac{b}{2} - \bar{y}_{Na} \right)^2 + E_3 \left( \frac{b}{2} + \bar{y}_{Na} \right)^2 \right) \tag{21}$$

$$K_{22} = EI_o + A_m \left[ E_1 \left( \frac{\sqrt{3}b}{3} - \bar{x}_{Na} \right)^2 + (E_2 + E_3) \left( \frac{\sqrt{3}b}{6} + \bar{x}_{Na} \right)^2 \right]$$

$$K_{12} = -A_m \left[ E_1 \bar{y}_{Na} \left( \frac{\sqrt{3}b}{3} - \bar{x}_{Na} \right) + \left( \frac{\sqrt{3}b}{6} + \bar{x}_{Na} \right) \left[ E_2 \left( \frac{b}{2} - \bar{y}_{Na} \right) - E_3 \left( \frac{b}{2} + \bar{y}_{Na} \right) \right] \right]$$

### *4.6. Boundary conditions*

For kinematic variables, the beginning of the manipulator is clamped and thus has known values of $\boldsymbol{p}(t,0) = \boldsymbol{p}_0$, $\boldsymbol{h}(t,0) = \boldsymbol{h}_0$ and $\boldsymbol{q}(t,0) = \boldsymbol{\omega}(t,0) = \boldsymbol{0}$. It is assumed that all actuators extend to the distal end of the robot with flat caps of area $A_i$ so that the magnitude of the force on the cap is $P_i A_i + E_i A_m v_{3i}^{Aa}$ which is the sum of actual pressure force and the equivalent RPE force in axial direction calculated in section 0. Therefore, the force and moment vectors applied from pneumatic pressure to end of neutral axis are determined using Eq. (22).

$$\boldsymbol{F}_i^b = \left( P_i A_i + E_i A_m v_{3i}^{Aa} \right) \boldsymbol{e}_3, \qquad \boldsymbol{L}_i^b = (\boldsymbol{r}_i - \boldsymbol{D}_{Na}) \times \boldsymbol{F}_i^b \tag{22}$$

Also, by introducing $m$ as the mass of the end cap, the weight is considered as a boundary condition.

$$\boldsymbol{F}^g = m\boldsymbol{g} \tag{23}$$

This force is applied to the central axis and for the homogeneous case where the central and neutral axes are identical, it has no moment. However, in the case of an inhomogeneous rod where the central and neutral axis are different, the moment from the weight would be obtained in the reference frame as Eq. (24).

$$\boldsymbol{L}^g = \boldsymbol{R}\boldsymbol{D}_{Na} \times \boldsymbol{F}^g \tag{24}$$

Finally, from the static equilibrium of the end-effector, the internal loads can be calculated.

$$\boldsymbol{n}(L) = \boldsymbol{F}^g + \boldsymbol{R}\sum_{i=1}^{3} \boldsymbol{F}_i^b, \qquad \boldsymbol{m}(L) = \boldsymbol{L}^g + \boldsymbol{R}\sum_{i=1}^{3} \boldsymbol{L}_i^b \tag{25}$$

### *4.7. Solving equations*

The constructed equations are a system of nonlinear, stiff, hyperbolic PDEs and contain derivatives of time $\boldsymbol{y}_t$ and arclength $\boldsymbol{y}_s$, which can be solved numerically by transferring them into discrete counterparts.

For semi-discretization in time, three types of backward differentiation formulas (BDF), including BDF1 (backward-Euler), BDF3 and BDF-$\alpha$, are employed. The general formula for a BDF can be written as Eq. (26) where $\boldsymbol{y}^h(t_i)$ is all the historical data.

$$\boldsymbol{y}_t(t_i) := c_0\boldsymbol{y}(t_i) + \boldsymbol{y}^h(t_i) \tag{26}$$

The unknown parameters of the proposed methods for a time step $dt$ are shown in Table 2.

Table 2. Parameters of different BDF methods

| | | |
|---|---|---|
| **BDF1** | $c_0 = \frac{1}{dt}$ | $\boldsymbol{y}^h(t_i) = c_1\boldsymbol{y}(t_{i-1})$ <br> where $c_1 = -\frac{1}{dt}$ |
| **BDF3** | $c_0 = \frac{11}{6dt}$ | $\boldsymbol{y}^h(t_i) = c_1\boldsymbol{y}(t_{i-1}) + c_2\boldsymbol{y}(t_{i-2}) + c_3\boldsymbol{y}(t_{i-3})$ <br> where $c_1 = -\frac{3}{dt}$, $c_2 = \frac{3}{2dt}$, $c_3 = -\frac{1}{3dt}$ |
| **BDF-α** | $c_0 = \frac{1.5+\alpha}{dt(1+\alpha)}$ | $\boldsymbol{y}^h(t_i) = c_1\boldsymbol{y}(t_{i-1}) + c_2\boldsymbol{y}(t_{i-2}) + d_1\boldsymbol{y}_t(t_{i-1})$ <br> where $c_1 = -\frac{2}{dt}$, $c_2 = \frac{0.5+\alpha}{dt(1+\alpha)}$, $d_1 = \frac{\alpha}{(1+\alpha)}$ |

For BDF-$\alpha$ method, the variable $\alpha \in [-0.5\ 0] \subset \mathbb{R}$ is a parameter that interpolates between $\alpha = -0.5$, named trapezoidal method and $\alpha = 0$, named BDF2 [30].

Using these approximations, by replacing all the time derivative terms according to Eq. (27), a set of ODEs is obtained in the spatial dimension that is only dependent on the solutions of previous ODEs.

$$\boldsymbol{v} = \boldsymbol{v}^* + \boldsymbol{K}_{se}^{-1}\boldsymbol{R}^T\boldsymbol{n}$$

$$\boldsymbol{u} = \boldsymbol{u}^* + \boldsymbol{K}_{bt}^{-1}\boldsymbol{R}^T\boldsymbol{m}$$

$$\boldsymbol{p}_s = \boldsymbol{R}\boldsymbol{v}$$

$$\boldsymbol{R}_s = \boldsymbol{R}\widehat{\boldsymbol{u}}$$

$$\boldsymbol{n}_s = -\boldsymbol{f}_e + \rho A\boldsymbol{R}(\Omega\boldsymbol{q} + \boldsymbol{q}^h) + \sum_{i=1}^{3}\left(P_iA_i + E_iA_m v_{3i}^{Aa}\right)\boldsymbol{R}_s\boldsymbol{e}_3 \quad (27)$$

$$\boldsymbol{m}_s = -\boldsymbol{l}_e - \boldsymbol{p}_s \times \boldsymbol{n} + \rho\boldsymbol{R}(\Omega\boldsymbol{J}\boldsymbol{\omega} + \boldsymbol{J}\boldsymbol{\omega}^h)$$
$$+ \sum_{i=1}^{3}\left(P_iA_i + E_iA_m v_{3i}^{Aa}\right)\boldsymbol{R}\left[\left(\boldsymbol{v} + \widehat{\boldsymbol{u}}(\boldsymbol{r}_i - \boldsymbol{D}_{Na})\right) \times \boldsymbol{e}_3 + (\boldsymbol{r}_i - \boldsymbol{D}_{Na}) \times \widehat{\boldsymbol{u}}\boldsymbol{e}_3\right]$$

$$\boldsymbol{q}_s = \Omega\boldsymbol{v} + \boldsymbol{v}^h - \widehat{\boldsymbol{u}}\boldsymbol{q}$$

$$\boldsymbol{\omega}_s = \Omega\boldsymbol{u} + \boldsymbol{u}^h$$

where $\Omega = \widehat{\boldsymbol{\omega}} + c_0\boldsymbol{I}$ and $E_iA_m v_{3i}^{Aa}$ is the equivalent RPE force in the axial direction.

This ODE system with known boundary conditions from section 0 leads to a boundary value problem. Such a BVP can be solved iteratively by guessing the unknown initial values as $\boldsymbol{G} = \{\boldsymbol{n}(t,0)\ \ \boldsymbol{m}(t,0)\}$ and spatially integrating using the forward Euler method or the classical fourth-order Runge-Kutta (RK4) method. The guessed values are then iteratively updated by a chosen nonlinear optimization routine in order to reduce the residual error of the distal boundary conditions to zero.

$$\boldsymbol{E}(\boldsymbol{G}) = \{\boldsymbol{E}^F\ \ \boldsymbol{E}^M\} \quad (28)$$

$$\boldsymbol{E}^F = \boldsymbol{n}(L) - \boldsymbol{F}^g - \boldsymbol{R}\sum_{i=1}^{n} \boldsymbol{F}_i^b - \boldsymbol{F}^e$$

$$\boldsymbol{E}^M = \boldsymbol{m}(L) - \boldsymbol{L}^g - \boldsymbol{R}\sum_{i=1}^{n} \boldsymbol{L}_i^b - \boldsymbol{L}^e$$

where $\boldsymbol{F}^e$ and $\boldsymbol{L}^e$ are, respectively, the external force and moment acting on the end of the neutral axis in reference frame.

This aspect of the shooting method can be implemented by a Levenberg-Marquardt algorithm with an adaptive damping coefficient as described in [39] or by a trust region dogleg method [40].

## 5. Experimental validation

### *5.1. Measurement setup*

The setup consists of three Festo SDE1 digital pressure sensors with an accuracy of 0.01-bar to measure the inlet pressure of the actuators. A 16-bar air compressor supplied the air for the actuators, and a 5/3-way Festo proportional solenoid valve, which works by a 10V control signal, controlled the actuator inlet air. The data from the pressure sensors was collected and the valve was operated using an Advantech PCI-1716HG data acquisition card. An Intel RealSense-SR300 digital camera was used to capture pictures of the actuator in different situations in order to derive the pose of the manipulator. The block diagram of the experimental setup and the relationship between the components is shown in Figure 7.

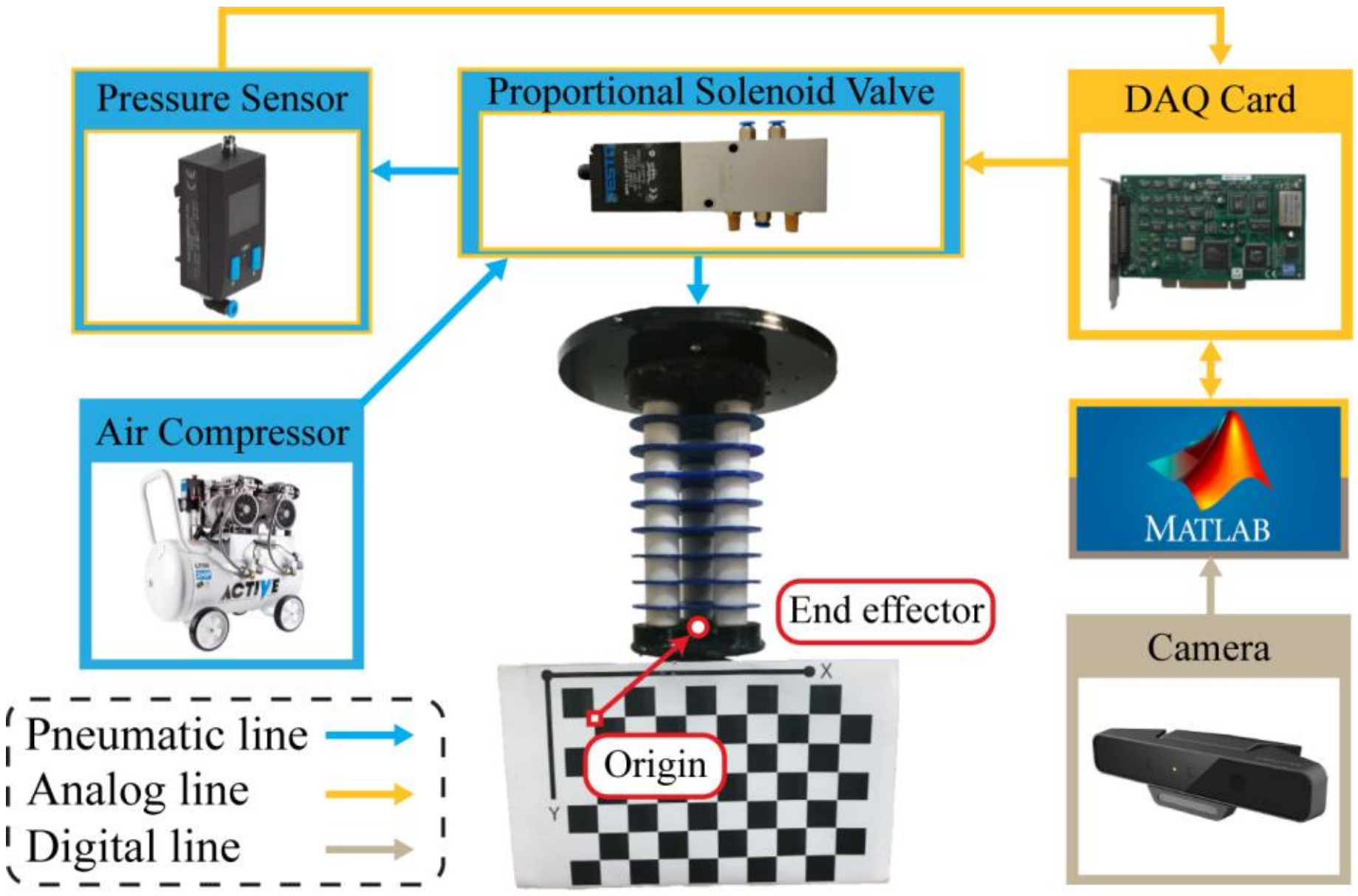

Figure 7. Experimental setup block diagram

A 3-D video reconstruction process based on direct linear transformation (DLT) is used to extract the pose of the manipulator end effector [41]. This is done by placing a checkerboard at the end of the manipulator. It is necessary to convert the extracted position from the origin of the checkerboard in order to obtain the exact position of the end-effector that is placed on the central axis. As mentioned in 0, for the case of an inhomogeneous rod, the neutral axis is different from the central axis, and therefore the calculated points from Cosserat equations that are in the framework of $f_{Na}$ must be transferred to the central framework ($f_C$) (Figure 8).

$$\{\boldsymbol{p}_C(s)\}_{f_C} = \{\boldsymbol{p}_{Na}(s)\}_{f_{Na}} + \boldsymbol{D}_{Na} - \boldsymbol{R}(s)\boldsymbol{D}_{Na} \qquad (29)$$

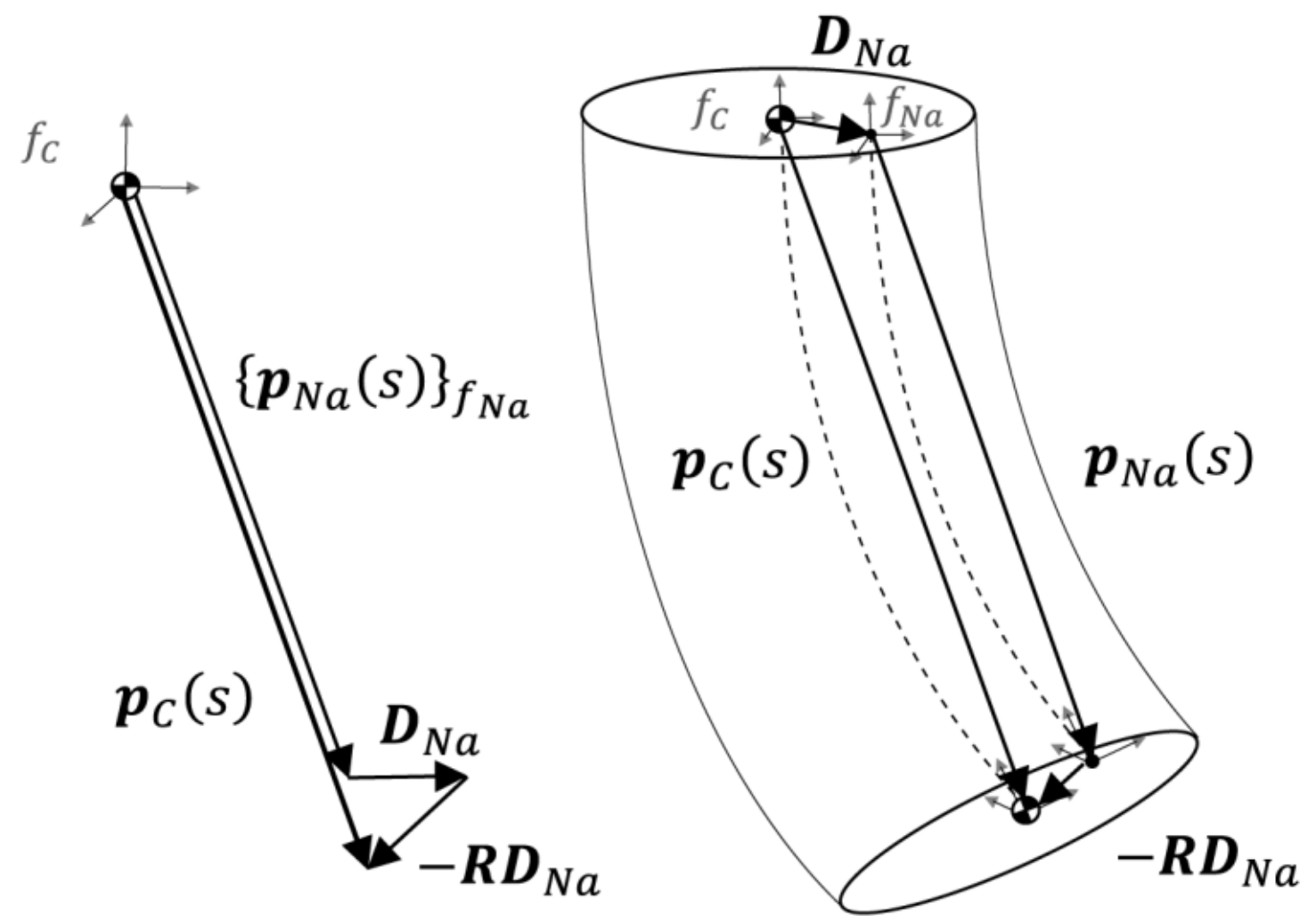

Figure 8. The relation between central and neutral axis

### *5.2. Model identification*

In this section, the necessary parameters are identified to solve Cosserat equations.

#### *5.2.1. RPE identification*

To solve the continuum model introduced in section 0, the exact magnitudes of $R_m$ and $\mu$ must be identified. Therefore, the elongation of actuators with $\Psi = 3°$ for two cases of under pressurization and external force is extracted. For this purpose, it is assumed that all three actuators are identical and are simultaneously pressurized from 0.15bar to 0.65bar with a 0.05bar step. External force is also applied by placing weights of 100g to 1200g with a step of 100g at the end of the robot. The unknown parameters are identified by using *fmincon* function of MATLAB and the results are presented in Figure 9.

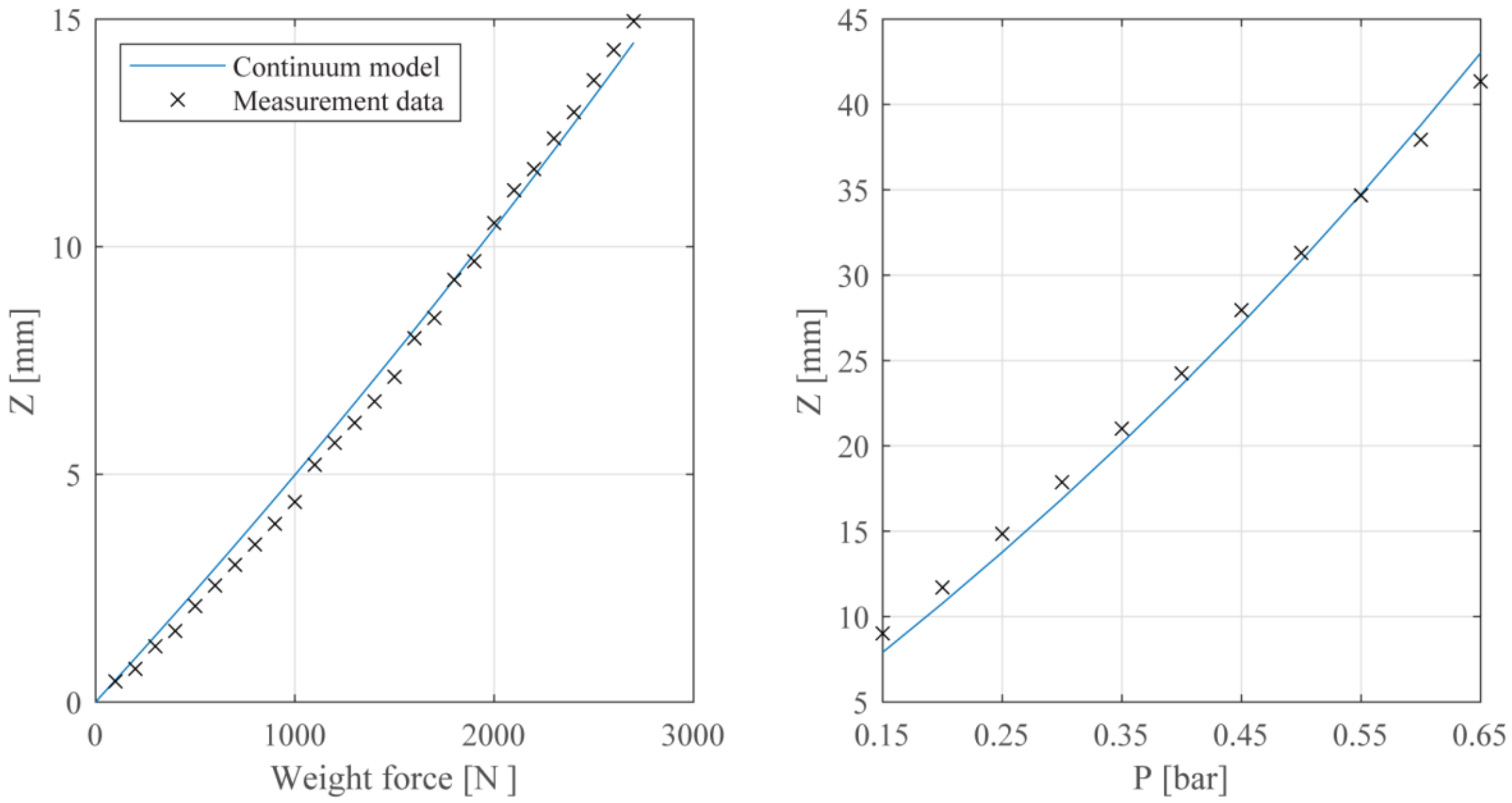

Figure 9. Actuator elongation due to external and pressure loads

After identifying the necessary parameters to solve the continuum model, it is now possible to find the elongation of actuators due to the radial pressure through Eqs. (6) and (14). This elongation can be approximated as a first-order polynomial in Eq. (30) using *polyfit*, a publicly available MALAB function, which fits with $R^2 = 0.99633$.

$$v_3^{Aa} = a * P + 1, \qquad a = 0.05324473 \tag{30}$$

*5.2.2. Material properties of Cosserat rod*

In Section 0, strains were introduced as a function of force and pressure. Therefore, if the actuators are examined without pressure and only under external forces, Eq. (31) can be achieved.

$$\boldsymbol{v} = \boldsymbol{v}(P = 0, \boldsymbol{n}(P = 0, \boldsymbol{F})) \ , \qquad \boldsymbol{u} = \boldsymbol{u}(P = 0, \boldsymbol{m}(P = 0, \boldsymbol{M})) \tag{31}$$

Using experimental data extracted in 0 with different weights, a relationship between force and strain for each actuator can be obtained. Therefore, in the above relation, $\boldsymbol{F} =$ $Fe_3$ and $\boldsymbol{M} = \mathbf{0}$. Then, by dividing the secondary length by the initial length $L_0$, the strains are obtained. Figure 10 shows the strain-force relationship.

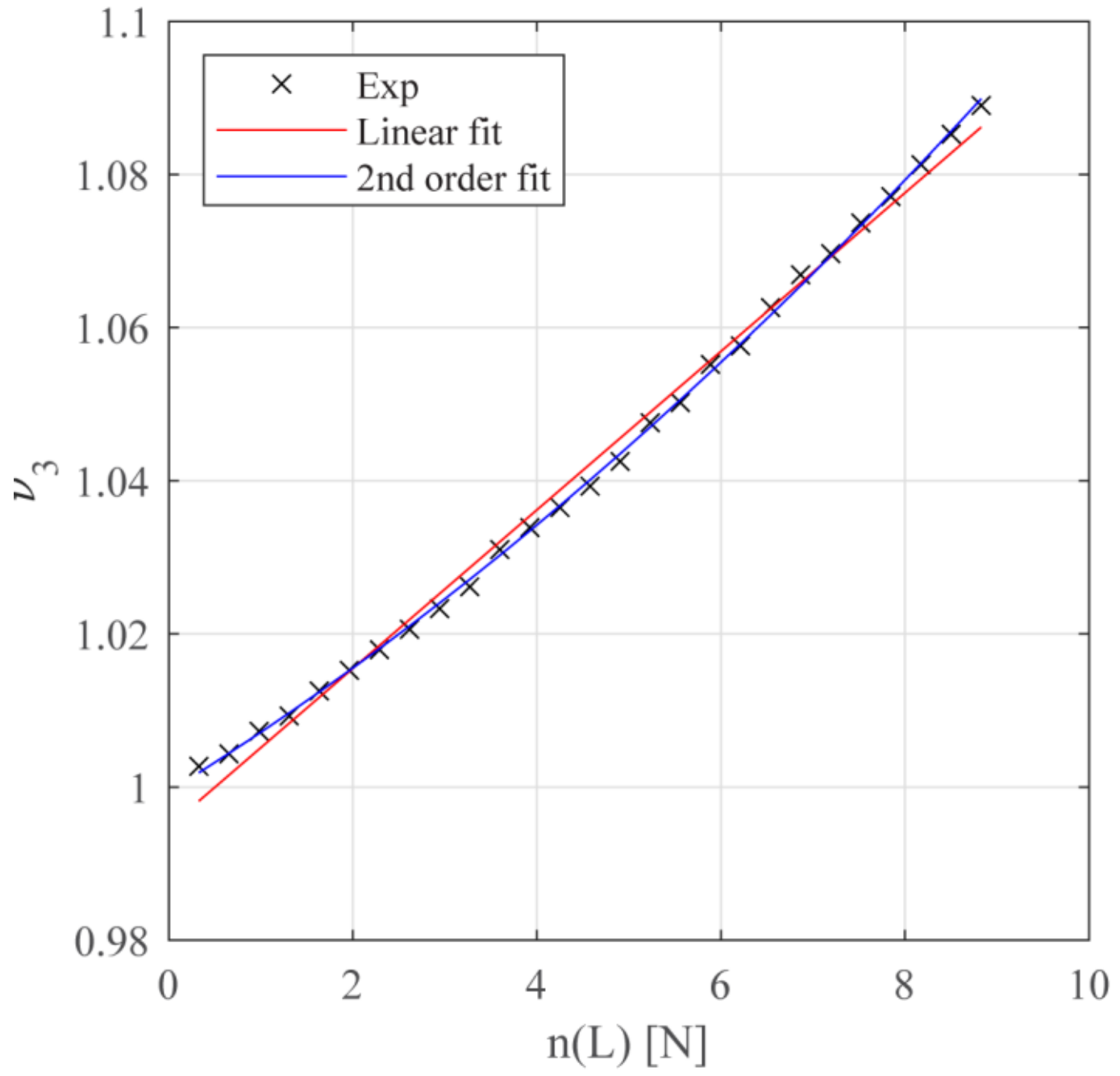

Figure 10. Strain-force diagram under external concentrated force and its approximations

If the resulted graph is approximated linear, by introducing $\beta$ as the slope of the graph, the Young's modulus can be calculated as $E_p = \frac{A_m}{\beta}$ which is constant and results in a homogeneous rod. In this case, the Young's modulus is $E$ = 289142.05$pa$.

On the other hand, if the graph is considered nonlinear, by changing the applied pressure on each actuator, the slope of the graph, which is related to the Young's modulus, will change. Therefore, the rod consisting of three actuators with different Young's modulus will be inhomogeneous. Here, this nonlinearity can be approximated as a second-order polynomial in Eq. (32) using *polyfit*, which fits with $R^2$ = 0.99943.

$$v(n) = a_2 n^2 + a_1 n + 1, \qquad a_2 = 0.00031962, \qquad a_1 = 0.00742681 \qquad (32)$$

Similar to the linear case, Eq. (33) calculates the Young's modulus from the slope of the graph.

$$E_i(n) = \frac{A_m}{\beta_i(n)} \quad \text{where} \quad \beta_i(n) = \frac{dv(n)}{dn} \tag{33}$$

In this way, the values of $E_1$ ‹$E_2$ and $E_3$ are obtained in terms of the axial force applied to each actuator and are calculated via the equilibrium of forces at the end cap. For example, in the case of pressurizing one actuator, the axial force on each of the other passive actuators is $n = \frac{PA_{in}}{2}$ and by that $E_i$ can be extracted from Eq. (33).

Furthermore, to identify the unknown parameters $f_s$ and $\nu$ for the shear modulus $G = \frac{E}{2f_s(1+\nu)}$, a new variable $\gamma = \frac{1}{2f_s(1+\nu)}$ is introduced, which is identified as $\gamma = 0.4094$.

## 6. Results

Experimental data is extracted for two cases: *Case (a)* actuating actuator 1 in Figure 6 and *Case (b)* actuating actuators 2 and 3 simultaneously. Both cases result in a planar movement, and therefore movements only in the direction of $x$ and $z$ are extracted. For static results, the end effector is captured by pressurizing the actuators from 0.15 bar to 0.65 bar with a step of 0.05 bar and the tool center point (TCP) position is shown in Figure 11.

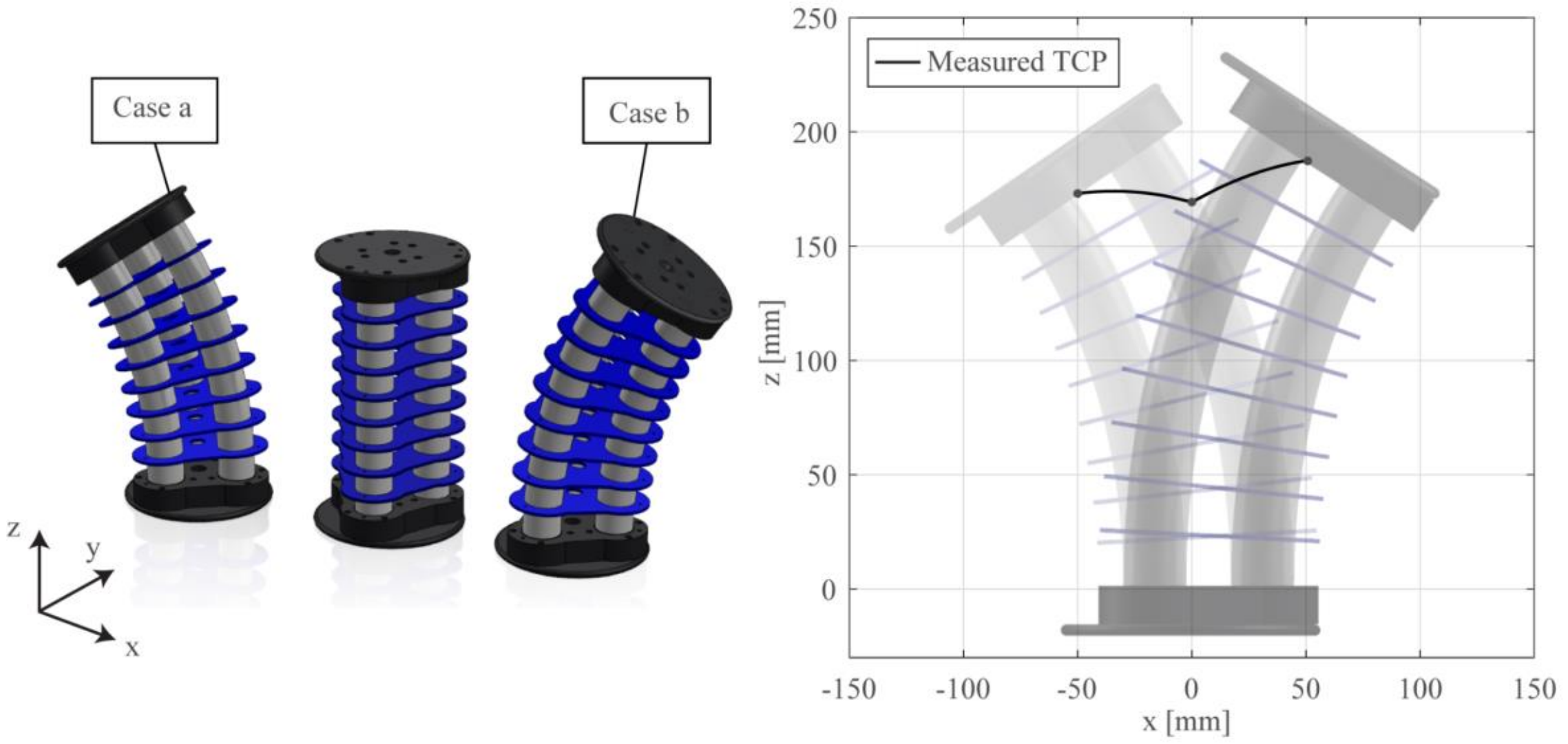

Figure 11. TCP trajectory for two introduced cases

Backbone deflection of the manipulator, assuming it as an inhomogeneous rod with RPE, is numerically calculated using RK4 method for spatial integration. It is compared with the experimental tip position in Figure 12.

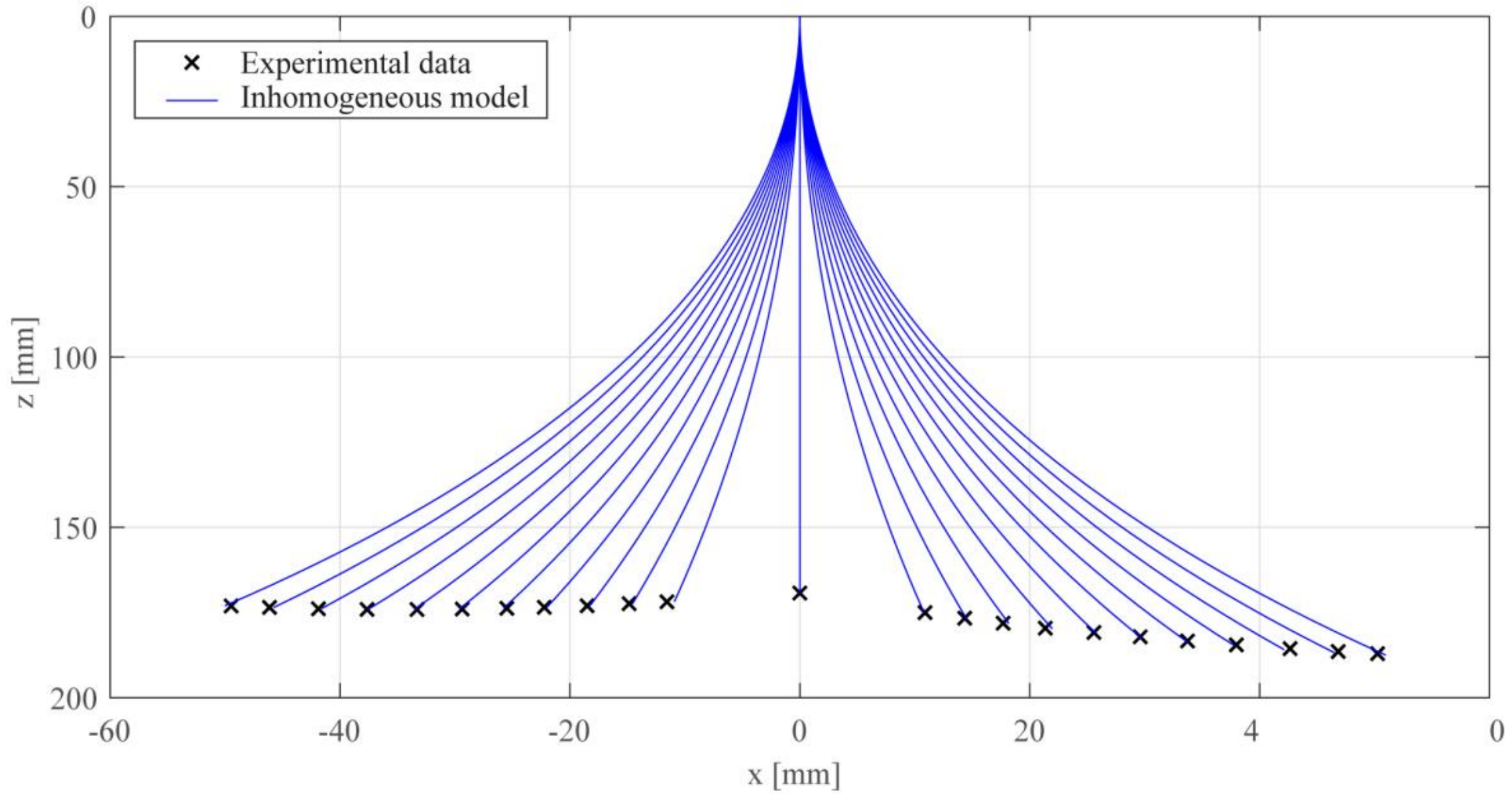

Figure 12. Comparison of the simulated backbone deflection and experimental TCP positions

In order to compare the accuracy of the model, the tip position and the normalized error in x and z directions are extracted for three simulations. The results are shown in Figure 13.

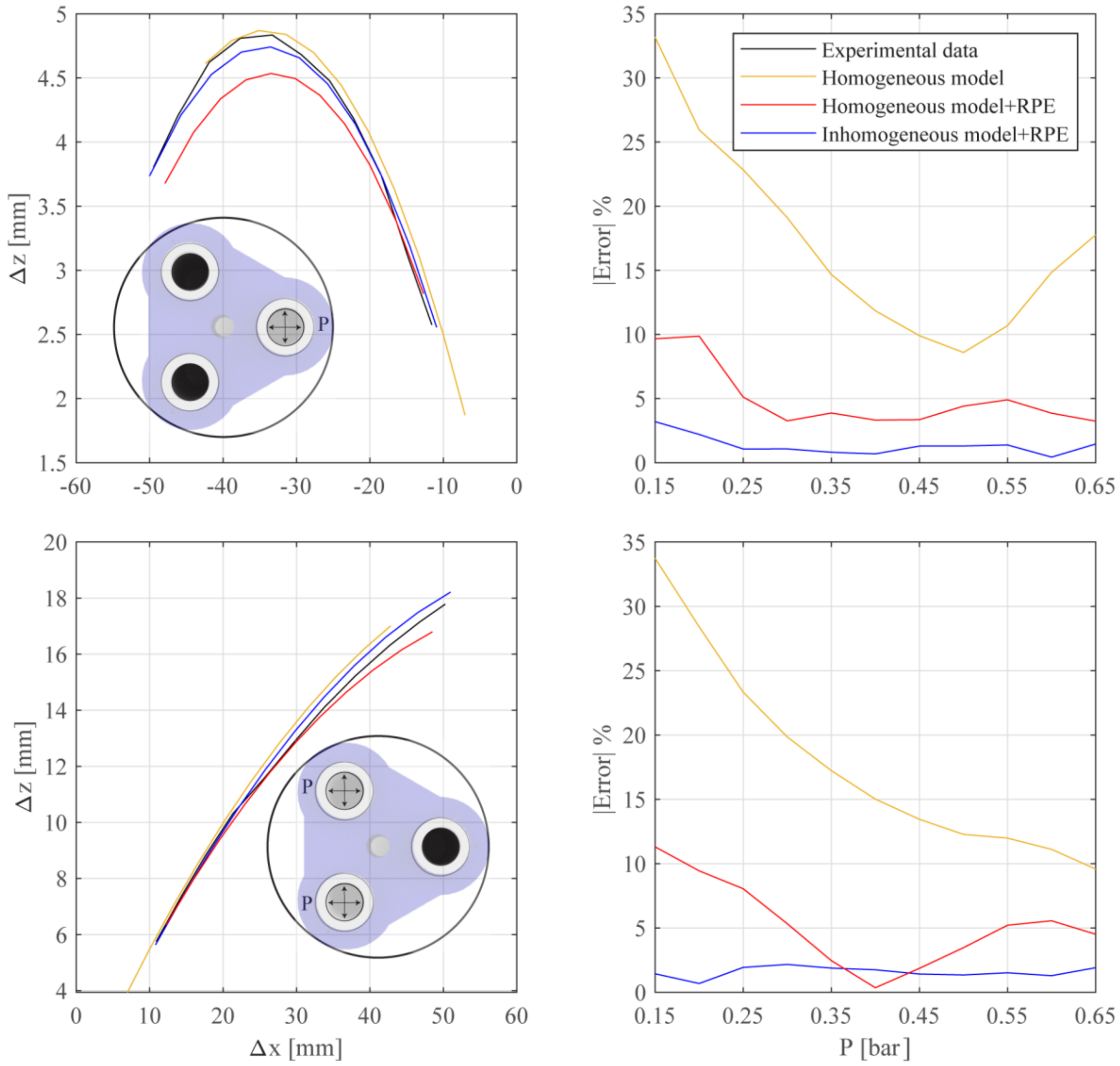

Figure 13. *left:* TCP positions and *right:* the normalized errors of simulated result

As Table 3 shows, the proposed method that assumes an inhomogeneous rod with RPE has improved the normalized error by 15.93%. On the other hand, considering the radial pressure alone causes 11.71% improvement.

Table 3. Mean value of normalized error for various cases

| **Inhomogeneity** | **Radial pressure effect** | **\|Error\|%** |
|---|---|---|
| ✕ | ✕ | 17.20 |
| ✕ | ✔ | 5.49 |
| ✔ | ✔ | 1.27 |

The extracted force-strain diagram in Figure 10 was almost linear, and therefore it is expected that the assumption of an inhomogeneous rod has little effect on the results. It should be noted that the inhomogeneity assumption will result in a greater improvement for an actuator with more nonlinear behavior, which for a fiber reinforced actuator can be achieved by increasing the angle of fibers [42].

To validate the dynamical model, a sinusoidal pressure $P_i = 0.35 + 0.25\sin\,(t + \frac{4\pi}{3} + \phi_0)$ is used as the input, and TCP of the manipulator is extracted. Furthermore, the dynamical equations of an inhomogeneous rod with RPE are solved using the proposed numerical methods in section 4.7. The performance of each method is compared in Table 4. The total simulated timeframe is 1 second with a timestep of 1/30 second, which is in accordance with the camera frame rate (30 fps). For the spatial integration, two cases of 50 and 200 discrete points are used.

Table 4. Comparison of numerical methods with Intel Core i7-7500U CPU @2.8GHz

| **Spatial integration method** | **Time discretization method** | **Runtime: Δt = 1/30 s Timeframe = 1s** | | **Average RMSE** | |
|---|---|---|---|---|---|
| | | **Spatial Steps = 50** | **Spatial Steps = 200** | **Spatial Steps = 50** | **Spatial Steps = 200** |
| Euler | Backward Euler | 0.5569 s | 1.7976 s | 1.2138 mm | 1.1343 mm |
| | BDF2 | 0.5615 s | 1.8442 s | 1.2183 mm | 1.1391 mm |
| | BDF-α, α = -0.1 | 0.5817 s | 1.8462 s | 1.2118 mm | 1.1322 mm |
| | BDF-α, α = -0.2 | 0.5810 s | 1.8538 s | 1.2087 mm | 1.1287 mm |
| | BDF-α, α = -0.3 | 0.6192 s | 2.2444 s | 1.2424 mm | 1.1613 mm |
| | Trapezoidal Method | Unstable | Unstable | Unstable | Unstable |
| | BDF3 | Unstable | Unstable | Unstable | Unstable |
| Fourth-Order Runge Kutta | Backward Euler | 1.6683 s | 6.4893 s | 1.1393 mm | 1.1395 mm |
| | BDF2 | 1.7196 s | 6.5137 s | 1.1724 mm | 1.1707 mm |
| | BDF-α, α = -0.1 | 1.6926 s | 6.5439 s | 1.1254 mm | 1.1259 mm |
| | BDF-α, α = -0.2 | 1.7942 s | 6.7770 s | 1.1221 mm | 1.1224 mm |
| | BDF-α, α = -0.3 | 2.0735 s | 7.7402 s | 1.2923 mm | 1.2817 mm |
| | Trapezoidal Method | Unstable | Unstable | Unstable | Unstable |
| | BDF3 | Unstable | Unstable | Unstable | Unstable |

As can be seen from Table 4, the Euler method for spatial integration with spatial steps of 50 leads to real-time solutions. Furthermore, the trapezoidal method and BDF3 need a higher time step to be stable.

The RK4 method for spatial integration along with BDF-$\alpha$ with $\alpha = -0.2$ for time discretization guides to the most accurate solution and thus is employed to validate the dynamical results in Figure 14.

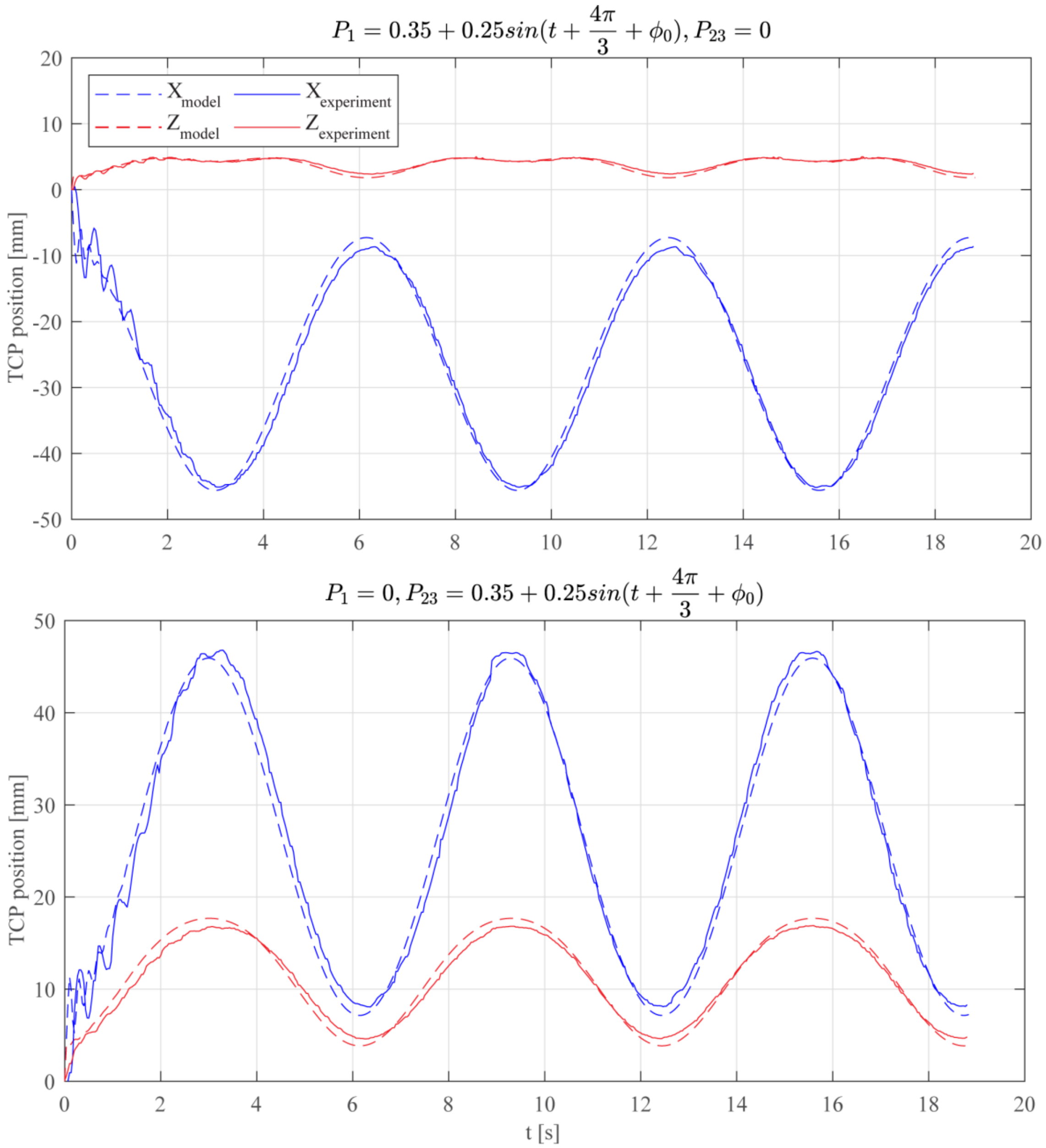

Figure 14. Dynamical results of sinusoidal pressure actuation

Table 5 shows the root mean square error in both directions between the proposed models and experimental data.

Table 5. Root mean square errors of dynamical results

| **Case** | **RMSE (mm)**<br>**Z direction** | **RMSE (mm)**<br>**X direction** |
|---|---|---|
| 1 | 0.3269 | 1.7349 |
| 2 | 0.8019 | 1.6247 |

## 7. Conclusion

In this study, a soft robotic manipulator with three coated fiber-reinforced actuators was introduced. First, a continuum method based on nonlinear elasticity technique was used to model the effect of radial pressure on the axial strains of a fiber-reinforced actuator. Then Cosserat rod theory was used to extract dynamical equations of the manipulator. Based on the behavior of the manipulator, two cases of homogeneous and inhomogeneous rods were implemented in constitutive equations, and the change of neutral axis was studied for the latter case. By adding the strains resulting from radial pressure that was calculated from the proposed continuum model, several methods were used to numerically solve the equations. From the experimental trials, the most accurate results were calculated using the RK4 method for spatial integration and BDF-$\alpha$ with $\alpha = -0.2$ for time discretization (with RMSE = 1.1221mm). Finally, solutions revealed that the proposed method was 15.93% more accurate than the conventional Cosserat rod method, in which the rod was homogeneous and the effect of radial pressure was not considered.

## Funding

The authors received no financial support for the research, authorship, and/or publication of this article.

## Disclosure statement

The authors report there are no competing interests to declare.

## ORCID

Sadegh Pourghasemi Hanza https://orcid.org/0000-0002-1521-1367

Hamed Ghafarirad https://orcid.org/0000-0001-7292-143X